\pdfoutput=1

\documentclass[11pt]{article}

\usepackage{authblk}
\usepackage[]{acl}

\usepackage{times}
\usepackage{latexsym}
\usepackage{amsmath}

\usepackage{algorithm}
\usepackage{algpseudocode}

\usepackage[T1]{fontenc}

\usepackage[utf8]{inputenc}

\usepackage{microtype}

\usepackage{url}

\usepackage{float}
\usepackage{graphicx}
\usepackage[caption=false]{subfig}

\usepackage{multirow}
\usepackage{enumitem}

\usepackage{amssymb}

\usepackage{color,soul}

\usepackage{todonotes}

\usepackage{footnote}
\usepackage{booktabs}
\usepackage{dsfont}

\newcommand*\samethanks[1][\value{footnote}]{\footnotemark[#1]}

\title{PreSumm: Predicting Summarization Performance Without Summarizing}

\author[1,2]{\bf Steven Koniaev\Thanks{~~Equal contribution.}}
\author[1,2]{\bf Ori Ernst\samethanks}
\author[1,2,3]{\bf Jackie Chi Kit Cheung}
{
\makeatletter
\renewcommand\AB@affilsepx{~~~~~~ \protect\Affilfont} \makeatother
\affil[1]{Mila – Quebec Artificial Intelligence Institute}
\affil[2]{McGill University}

\affil[3]{Canada CIFAR AI Chair, Mila}
}
\affil[   ]{} 

\affil[  ]{\tt \{oriern\}@gmail.com}
\affil[  ]{\tt \{steven.koniaev@mail., jackie.cheung@\}mcgill.ca}

\date{}

\begin{document}
\maketitle

\begin{abstract}
Despite recent advancements in automatic summarization, state-of-the-art models do not summarize all documents equally well, raising the question: why? While prior research has extensively analyzed summarization models, little attention has been given to the role of document characteristics in influencing summarization performance.
In this work, we explore two key research questions. First, do documents exhibit consistent summarization quality across multiple systems? If so, can we predict a document’s summarization performance without generating a summary? We answer both questions affirmatively and introduce PreSumm, a novel task in which a system predicts summarization performance based solely on the source document. Our analysis sheds light on common properties of documents with low PreSumm scores, revealing that they often suffer from coherence issues, complex content, or a lack of a clear main theme.
In addition, we demonstrate PreSumm’s practical utility in two key applications: improving hybrid summarization workflows by identifying documents that require manual summarization and enhancing dataset quality by filtering outliers and noisy documents.
Overall, our findings highlight the critical role of document properties in summarization performance and offer insights into the limitations of current systems that could serve as the basis for future improvements.

\end{abstract}

\section{Introduction}
\label{sec_intro}
Recent years have witnessed a remarkable proliferation of summarization models, with many achieving impressive performance on widely used benchmarks. These models, often rooted in large-scale language modeling, represent a significant leap forward in natural language processing capabilities.

\begin{figure}[t]
    \centering
    \resizebox{1.05\linewidth}{!}{
    \includegraphics{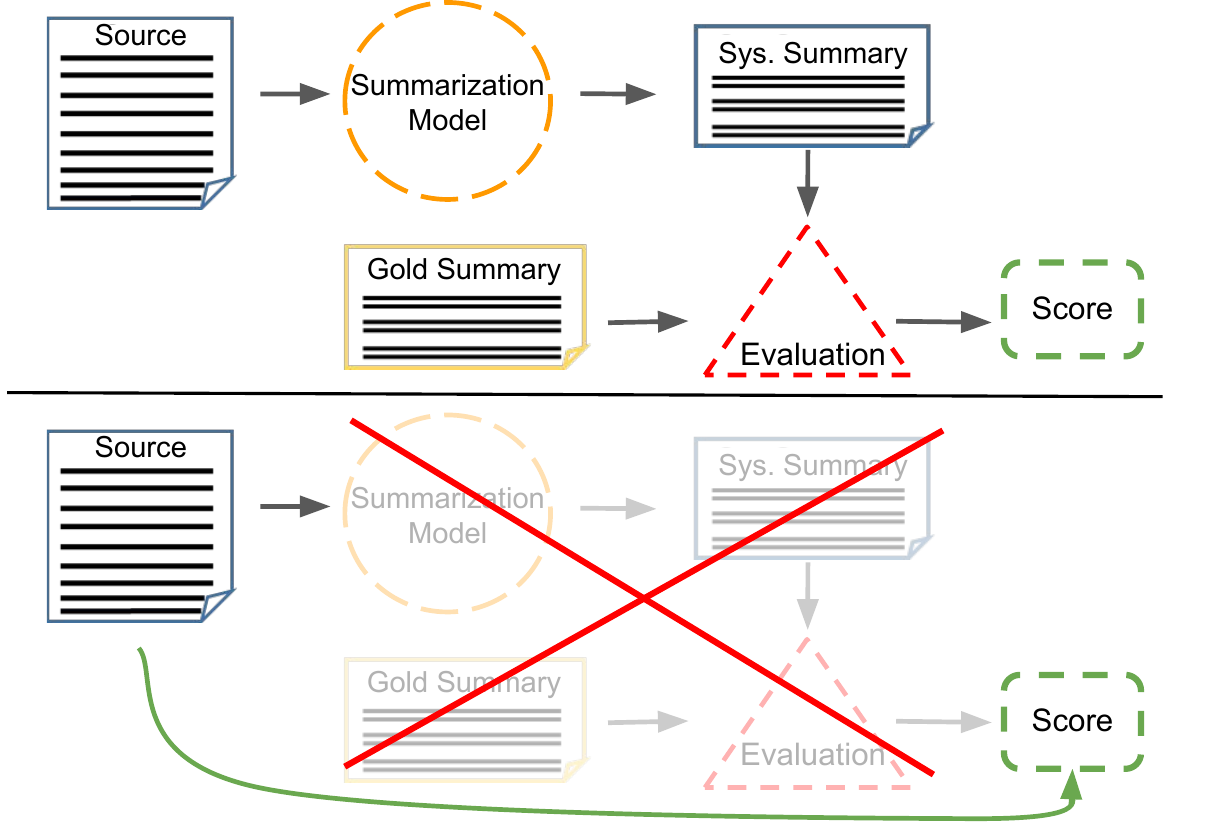}}
    \caption{An illustration comparing the traditional evaluation process (top) with our approach (bottom).}
\label{fig:illustration}
\end{figure}

Despite these advancements, not all documents are equally well summarized by these models. Some documents yield better summaries, while others do not. As many summarization models share inherent design principles, operational mechanisms, and therefore limitations, we hypothesize that documents that are `difficult' for one system to summarize tend to also be `difficult' for other systems. If this hypothesis holds, it suggests that certain intrinsic properties of source documents systematically hinder summarization performance. While prior work has extensively analyzed the characteristics of summarization \textit{models} \citep{Goyal2022NewsSA, zhang-etal-2024-benchmarking}, there is a lack of research on how \textit{document} properties influence performance across systems.

To bridge this gap, we introduce the PreSumm task. In this task a system should predict a document’s performance in future summarization tasks based only on the document, without generating a summary. An illustration of our approach is presented in Figure \ref{fig:illustration}. Success in this task suggests that we can distinguish in advance documents that most models summarize successfully, from those where performance falters. 

Analyzing such a successful PreSumm model can reveal critical source-based features that challenge current systems. Gaining deeper insights into these document-level limitations could not only allow targeted improvements in summarization models but also enable strategic pre-processing document edits to facilitate the summarization process.

Additionally, PreSumm offers several practical benefits that could improve the efficiency and effectiveness of summarization workflows.
One key advantage is its potential for hybrid systems where humans can focus their attention on more difficult summary cases. For example, organizations often evaluate their output by relying on automatic metrics or manual human evaluation, which can reveal poor-quality summaries that require further manual summarization—an expensive and time-consuming process. By leveraging PreSumm to identify low-performing documents in advance, organizations can prioritize these cases for manual summarization or decide to opt-out before engaging in costly summarization workflows, thereby saving valuable resources and optimizing operational efficiency.

Furthermore, a PreSumm model may identify outliers and noisy documents in the dataset whose removal could enhance outcomes. An appealing example is multi-document summarization task, where one or more problematic documents might degrade the overall quality of the output. By filtering out such documents, PreSumm can help ensure more consistent, high-quality results across summarization tasks. These practical applications highlight PreSumm’s potential as a valuable tool for improving both the cost-effectiveness and performance of summarization systems.

To enable this approach, we first confirm our initial hypothesis by demonstrating a moderate correlation between document performance ranking across various systems (Section \ref{subsec_preliminary_analysis}). Next, we train several models for the PreSumm task, relying solely on the source document (Section \ref{sec_experiments}).
Surprisingly, the supervised PreSumm models exhibit a stronger correlation with human evaluations not only compared to other baselines but also surpassing conventional automatic metrics, despite these metrics having access to the system summary. These findings also support our hypothesis that certain global document features influence summarization performance across different systems.

We also show promising results in two downstream tasks as an extrinsic evaluation. First, leveraging PreSumm predictions outperforms baselines in detecting documents that require manual summarization (Sec. \ref{sec_task_filtering}). Second, we improve summarization of multi document sets by filtering out documents with low PreSumm scores (Sec. \ref{sec_task_MDS}). 

A deeper analysis reveals that our best PreSumm model assigns low scores to documents with coherence problems, complex content, and without a clear main theme (Section \ref{sec_analysis}).
Overall, to the best of our knowledge, we are the first to focus on document properties that hinder modern summarization systems. By providing deeper insights into these limitations, PreSumm establishes a strong foundation for targeted advancements and future improvements in summarization research.\footnote{Our best model and its predictions over test data is publicly available at \href{https://github.com/oriern/PreSumm}{https://github.com/oriern/PreSumm}.}

\section{Related Work}
Our work draws significant inspiration from PreQuEL \citep{don-yehiya-etal-2022-prequel}, which introduces a similar approach centered on machine translation. PreQuEL aims to predict translation system performance based solely on the source text. While our motivation and general methodology align with theirs, to the best of our knowledge, we are the first to apply this approach to text summarization. By leveraging PreSumm, we have enhanced summarization-specific applications and explored features more relevant to summarization, making our contributions distinct and novel.

While recent approaches (e.g., \citep{vig-etal-2022-exploring, zhang-etal-2024-benchmarking} focused on embedding representation and did not leverage explicit source-based features, in the past, many summarization systems relied on explicit document-based features. These systems focused primarily on selecting key source sentences for inclusion in the summary—a task often framed as a classification problem. These features ranged from the presence of cue phrases \citep{Gupta2011SummarizingTB, 10.3844/jcssp.2010.1366.1376}, the inclusion of numerical data \citep{prasad2012feature, Abuobieda2012TextSF}, sentence length \citep{FATTAH2009126, Abuobieda2012TextSF}, and sentence position \citep{10.1007/978-3-642-28601-8_31, FATTAH2009126, Abuobieda2012TextSF, li-etal-2016-role}, to discourse structure \citep{louis-etal-2010-discourse}, among others. While these studies employed such features to generate summaries, one may suggest a potential link between these features and the performance of summarization models. However, this property was not explicitly examined. In contrast, our work explores \textit{system-independent} \textit{document-level} features that impact the performance of modern summarization models in both \textit{abstractive} and \textit{extractive} modes.

\section{Task Definition}
\label{sec_taskDefinition}
Our task is to predict the average performance of summarization systems on a given document, using only the document as input. Averaging performance across multiple systems can reveal key properties of the document itself, while minimizing the influence of system-specific variability or noise.
Formally, let $\mathcal{D}$ represent a corpus of $N$ text passages, where each document is denoted as $d_i$. PreSumm aims to estimate the average quality of summaries generated by multiple systems for each document.

Specifically, consider $M$ systems, denoted as $s_1, \dots, s_M$, where system $s_i$ produces a summary for document $d_j$ that is scored as $S_{i,j}$. The goal of PreSumm is to first be able to generate a function $f : \mathcal{D} \rightarrow \mathbb{R}$ that predicts the average score assigned to the summaries of document $d_j$ across all systems:
\begin{equation}
    d_j^* = \frac{1}{M}\sum_i^M S_{i,j}
\end{equation}

This score, $d_j^*$, reflects the average quality of the summaries generated by different systems for document $d_j$. We could leverage this score to rank the documents by their potential performance.

To evaluate the model's performance on ranking documents, we measure the correlation between the predicted scores and the gold-standard scores, following standard practices for assessing summarization metrics \citep{fabbri-etal-2021-summeval}.

\section{Dataset and Preliminary Analysis}
\label{sec_data}

\subsection{The RoSE Dataset}
The RoSE dataset \citep{liu-etal-2023-revisiting} introduces a new method for annotating summarization datasets, which improves annotator agreement via \textit{Atomic Content Units} (ACUs). The protocol tasks an annotator to convert a reference summary into atomic factual statements, then to compare the generated summary to these ACUs. The ACU score is defined as $f(s,A)=\frac{|A_s|}{|A|}$ where $|A|$ is the total number of ACUs for a given reference summary and $|A_s|$ is the number of matched ACUs of the generated summary with respect to the reference.

The authors manually evaluated over 22,000 summary-level annotations across 2,500 documents summarized by 28 top-performing systems on three datasets (CNN/DailyMail \citep{nallapati-etal-2016-abstractive}, XSum \citep{narayan-etal-2018-dont}, SamSum \citep{gliwa-etal-2019-samsum}). This extensive manual evaluation of diverse systems and datasets aligns well with our task, and we adopt the ACU score as the gold score $S_{i,j}$ to predict. 
For our test set, we uniformly sampled 20\% of the documents from each dataset in RoSE, totaling 500 documents. The remaining 2,000 documents were used for the training set. Data statistics can be found in Table \ref{fig:data_stat}.

\begin{table}[t]
\begin{center}
\resizebox{0.7\columnwidth}{!}{
\begin{tabular}{l c c c} 

 \toprule
 Dataset  & \#Doc & \#Sys. & \#ACU \\ 
 \midrule
 CNNDM    & 1,500 & 8-12& 17.2k  \\
 XSum & 500 &8 & 2.3k \\
 SamSum & 500&8& 2.3k \\
 \bottomrule
\end{tabular}
}
    \caption{Distribution of RoSE Dataset. Taken from \citet{liu-etal-2023-revisiting}. }
    \label{fig:data_stat}
    \end{center}
\end{table}

\subsection{Preliminary Analysis - Do Systems Fail on The Same Documents?}
\label{subsec_preliminary_analysis}

In this section, we investigate whether different systems tend to consistently fail or succeed across the same documents. Since each system generates summaries for the same set of documents, their performance scores can be used to rank the documents. Our hypothesis is that different systems rank documents similarly, meaning the systems on certain documents consistently perform well or poorly across various systems.

To test this hypothesis, we measure the correlation between the ACU scores of summaries generated by different systems for the same document. Specifically, we calculate the correlation between the ACU scores of systems $l$ and $k$ as $corr(S_{l,1..N}, S_{k,1..N})$. We then compute the average correlation across all systems using the formula:
\begin{equation}
    \frac{2}{M^2-M}\sum_{l=1}^M \sum_{k=1}^{l-1} corr(S_{l,1..N},S_{k,1..N})
\end{equation}

We obtained a Kendall Tau correlation of 0.446 and a Spearman correlation of 0.565 over the entire RoSE dataset, indicating a moderate agreement in document rankings across systems. This suggests that many documents retain their relative ranking, regardless of the system used for summarization. These findings support our assumption that certain document-specific features significantly influence summarization performance. As a result, it might be possible to predict a document's average summarization performance based solely on its intrinsic characteristics.

\section{Experiments}
\label{sec_experiments}
We examine the ability of several methods to rank the documents according to the average score across all systems,  $d_j^* = \frac{1}{M}\sum_i^M S_{i,j}$. In this section we present simple baseline models based on document-based features (Section \ref{subsec_baseline_models}) and several supervised models trained on our training data (Section \ref{subsec_supervised_models}). In addition, we compare these methods to conventional summarization automatic metrics that utilize not only the document itself, but also the generated summary or even a reference summary (Section \ref{subsec_baseline_models_automatic_metrics}).

\subsection{Baseline Models}
\label{subsec_baseline_models}

\paragraph{Document Statistics.} We explored several basic statistics about the documents, including document length by word, the number of numerical values, and the number of unique named entities identified using the NER module from the NLTK package. All of these features might be associated to the reading complexity of documents, where longer documents with more numeric details and name entities might be more complex to read.

\paragraph{Flesch–Kincaid Readability Tests.} We applied the Flesch Reading Ease test \cite{flesch1948new} to measure how easy the document is to read, with scores ranging from 1 to 100  where higher scores indicate easier readability. 
Similarly, we used the Flesch-Kincaid Grade Level to estimate the U.S. education grade level required to understand the text, with higher scores corresponding to a more advanced reading level.

\subsection{Supervised Models}
\label{subsec_supervised_models}

We trained several models with a fixed number of 5 epochs over our training data. More implementation details are elaborated in Appendix \ref{subsec_Appendix_implement}.

\paragraph{Regression.} In this model, we trained the model to predict the actual $d_j^*$ scores. The input is a document $j$ and the target is $d_j^*$. We leveraged the Longformer model \cite{DBLP:journals/corr/abs-2004-05150} to allow a large input length of 4096 tokens required in many documents from our set. We added a regression head on top of the output of the 784 dimensional [CLS] token of the last layer of the Longformer. The regression head contains a feed forward component with an output of one dimension, which should predict the $d_j^*$ score. Additionally, we used a standard MSE loss function for training. We denote this model as PreSummReg.

\paragraph{Classification.}
Since some applications may only care about document rankings rather than their exact $d_j^*$ scores, we train an additional model to rank the documents directly based on pairwise comparisons. This approach aligns with our evaluation metric, which is based on correlation. Instead of predicting individual scores, we aim to determine which document in a pair should be ranked higher, and then aggregate these local decisions to form a global ranking.

In this model, the input consists of a pair of documents, and the task is to classify whether the first document should be ranked higher than the second. We represent each document using embeddings from a pre-trained Longformer model, concatenate the two embeddings, and feed the result into a RankNet model \cite{burges2005learning}. The target label for each pair of documents $(i, j)$ is $\delta_{ij} = \mathds{1}_{d_i^* > d_j^*}$, where $\mathds{1}$ is an indicator function that equals 1 if the condition $d_i^* > d_j^*$ is fulfilled, or 0 otherwise. The model is trained using a binary cross-entropy loss function.

We generate all possible $n^2$ pairs from the training set to fine-tune the model and use all $m^2$ pairs from the test set during evaluation. To derive the final global ranking, we follow the method outlined by \citet{keswani-jhamtani-2021-formulating}, where the final score for document $i$ is defined as $S(i) = \sum_j \hat{\delta}{ij}$, with $\hat{\delta}{ij}$ representing the predicted outcome for the document pair $(i, j)$. We then sorted $S(i)$ scores for the final ranking. We denote this model as PreSummClas.

\paragraph{Frozen Weights.}
Given the relatively small amount of training data, we explored a model with fewer trainable parameters to better handle the limited dataset. Specifically, we employed a variant of the regression model, freezing all weights except those in the regression head. We denote this model as PreSummRegFroz.

\subsection{Summarization Automatic Metrics}
\label{subsec_baseline_models_automatic_metrics}

While our approach focuses on methods that rely solely on the document, we compare them to automatic summarization metrics that do not share this limitation. Specifically, we compare against two reference-free metrics, BLANC \citep{vasilyev-etal-2020-fill} and SummaQA \citep{scialom-etal-2019-answers}, which use both the document and the system-generated summary, as well as two reference-based metrics, ROUGE \citep{lin-2004-rouge} and BERTScore \citep{zhang2019bertscore}, which evaluate the system summary against a reference summary. To obtain a per-document score using these metrics, we averaged the metric scores assigned to all summaries generated from the same document.

\subsection{Results}

\begin{table}[t!]
\begin{center}
\resizebox{\columnwidth}{!}{
\begin{tabular}{l c c c} 
 \toprule
 Method & Kendall $\tau$  & Pearson \textit{r} & Spearman\\ 
  \midrule
 
 Document length & -0.005 & -0.048 & -0.010\\ 
 Count of Numbers &  -0.016   & -0.106 & -0.023 \\
  \# Unique Named Entities & -0.054 & -0.134 & -0.071\\
 Flesch Reading Ease & -0.016 & 0.030 & -0.021 \\
 Flesch Kincaid Grade & -0.0489 & -0.0483 & -0.0104 \\

 PreSummReg (Ours) & \textbf{0.321}   & \textbf{0.463} & \textbf{0.460} \\ 
 PreSummClas (Ours) &  0.306 &0.389& 0.389\\
  PreSummRegFroz (Ours) & 0.279 & 0.406 & 0.403 \\
 \midrule
  Blanc & 0.246 & 0.322 & 0.347 \\
 SummaQA & 0.167 & 0.241 & 0.286 \\
 ROUGE & 0.431 & 0.652 & 0.597 \\
 BERTScore & 0.252 & 0.430 & 0.351 \\

\bottomrule
\end{tabular}
}
    \caption{Test set correlations of different PreSumm methods. The top methods use the document only, while the bottom section of methods also uses the system summary.}
    \label{tab:PreSummResults}
   
    \end{center}
     
\end{table}

All models are evaluated by measuring their correlation with the gold-standard ACU labels, which reflects how well they ranked the documents. The correlation scores are summarized in Table \ref{tab:PreSummResults}. Most baselines show near-zero correlation, except for the number of numeric values and named entities, which exhibit a small negative correlation. This suggests that an increased presence of numbers and entities may lead to lower summarization model performance. In contrast, most trained models achieve a moderate correlation, supporting our hypothesis that document ranking can, to some extent, be predicted based solely on the document content.

The table shows that our supervised models outperform all baseline models. Notably, they even surpass both reference-free metrics, despite these metrics utilizing the system summary, as well as BERTScore, which leverages both the system and reference summaries—whereas PreSummReg relies solely on the document. Although ROUGE outperforms our supervised models, it also benefits from access to both the system and reference summaries. These comparisons highlight the strong performance of our supervised models.
Given its superior performance and simplicity, we selected the PreSummReg model to represent the PreSumm approach in the subsequent experiments.

\subsection{Out-of-Distribution Performance}
\label{subsec_OOD}

\begin{table}[t]
\begin{center}
\resizebox{0.6\columnwidth}{!}{
\begin{tabular}{l c c} 

 \toprule
 Dataset  & \#Topics & \#Sys.  \\ 
 \midrule
 DUC 2006    & 20 & 22  \\
 DUC 2007    & 23 & 13  \\
 TAC 2008 & 48 & 57 \\
 TAC 2009 & 44&55 \\
 TAC 2010 & 44&55 \\
 TAC 2011 & 44&50 \\
 \bottomrule
\end{tabular}
}
    \caption{Number of topics (document sets) and system summaries that participated in Pyramid evaluation by dataset}
    \label{tab:pyr_stat}
    \end{center}
\end{table}

\begin{table*}[t]
    \centering
    \resizebox{\linewidth}{!}{
    \begin{tabular}{lcccccccccccccccccc}
        \toprule
        \multirow{2}{*}{Dataset}& \multicolumn{3}{c}{DUC 2006} & \multicolumn{3}{c}{DUC 2007} & \multicolumn{3}{c}{TAC 2008} & \multicolumn{3}{c}{TAC 2009} & \multicolumn{3}{c}{TAC 2010} & \multicolumn{3}{c}{TAC 2011} \\
        \cmidrule(lr){2-4} \cmidrule(lr){5-7} \cmidrule(lr){8-10} \cmidrule(lr){11-13} \cmidrule(lr){14-16} \cmidrule(lr){17-19} 
        & K  & P & S& K  & P & S& K  & P & S& K  & P & S& K  & P & S & K  & P & S \\
        \midrule
        Length &\textbf{-0.26} & -0.34 & \textbf{-0.39} & 0.00 & 0.10 & 0.00 & 0.01 & 0.03 & 0.02 & 0.15 & 0.14 & 0.20 & -0.02 & -0.11 & -0.02 & -0.10 & -0.07 & -0.16 \\
        \# Unique NEs & -0.25 & -0.27 & -0.37 & 0.06 & 0.27 & 0.12 & -0.01 & -0.01 & 0.01 & \textbf{0.23} & 0.25 & \textbf{0.36} & 0.07 & 0.08 & 0.13 & 0.16 & 0.25 & 0.19 \\
        Flesch & 0.05 & 0.30 & 0.15 & 0.04 & 0.06 & 0.08 & -0.03 & -0.05 & -0.03 & 0.18 & 0.24 & 0.25 & 0.03 & -0.01 & 0.03 & 0.28 & 0.37 & 0.42 \\

       PreSummReg & 0.25 & \textbf{0.39} & 0.37 & \textbf{0.31} & \textbf{0.43} & \textbf{0.46} & \textbf{0.13} & \textbf{0.21} & \textbf{0.19} & 0.13 & \textbf{0.27} & 0.21 & \textbf{0.25} & \textbf{0.43} & \textbf{0.39} & \textbf{0.40} & \textbf{0.44}& \textbf{0.60} \\
        \bottomrule
    \end{tabular}
    }
    \caption{Correlation scores (Kendall-tau [K], Pearson [P], Spearman[S]) of various methods to different Pyramid dataset scores of \textit{system} summaries}
    \label{tab:correlation_results_pyr}
\end{table*}

\begin{table*}[t]
    \centering
    \resizebox{0.9\linewidth}{!}{
    \begin{tabular}{lcccccccccccc}
        \toprule
        \multirow{2}{*}{Dataset} &  \multicolumn{3}{c}{TAC 2008} & \multicolumn{3}{c}{TAC 2009} & \multicolumn{3}{c}{TAC 2010} & \multicolumn{3}{c}{TAC 2011} \\
        \cmidrule(lr){2-4} \cmidrule(lr){5-7} \cmidrule(lr){8-10} \cmidrule(lr){11-13}
        & K  & P & S & K  & P & S& K  & P & S & K  & P & S\\
        \midrule
        Length  &\textbf{-0.15} & -0.30 & \textbf{-0.22} & -0.06 & -0.05 & -0.06 & -0.07 & -0.14 & -0.11 & -0.22 & -0.17 & -0.31 \\
        
        \# Unique NEs & -0.14 & \textbf{-0.31} & -0.21 & -0.02 & -0.07 & -0.03 & -0.05 & -0.04 & -0.07 & -0.05 & 0.04 & -0.07 \\
        
        Flesch & -0.05 & -0.08 & -0.10 & 0.03 & \textbf{0.14} & 0.06 & 0.02 & 0.12 & 0.03 & 0.23 & 0.33 & 0.31 \\
        
        PreSummReg & 0.04 & 0.09 & 0.05 & \textbf{0.09} & \textbf{0.14} & \textbf{0.12} &\textbf{0.15} & \textbf{0.23} & \textbf{0.22} & \textbf{0.32} & \textbf{0.47} & \textbf{0.46} \\

        \bottomrule
    \end{tabular}
    }
    \caption{Correlation scores (Kendall-tau [K], Pearson [P], Spearman[S]) of various methods to different Pyramid dataset scores of \textit{reference} summaries}
    \label{tab:correlation_results_pyr_refSumm}
\end{table*}

Our PreSummReg model was trained and evaluated on specific models and datasets. In this section, we investigate its ability to generalize to other datasets with systems not seen during training.

To that end, we adopted the Pyramid human annotations \citep{nenkova-passonneau-2004-evaluating}, which have been applied over several years to DUC and TAC multi-document summarization benchmarks, involving approximately 50 different systems per year (details in Table \ref{tab:pyr_stat}). To the best of our knowledge, this is the only resource for human evaluation of automatic summarization that does not involve the datasets we have used for training PreSummReg. Thus, it serves as an ideal benchmark to assess generalization.

We measured the correlation of PreSummReg to the Pyramid scores, and compared the performance against several baselines. To apply our single-document-based model and other baselines to the multi-document setup, we calculated scores for each document individually and then averaged them across each document set. As presented in Table \ref{tab:correlation_results_pyr}, PreSummReg consistently outperforms the baselines across most datasets by a large margin, demonstrating its strong generalization capability to unseen models and datasets.

In addition to \textit{system} performance, the Pyramid annotation method also evaluates the quality of \textit{reference} summaries in some datasets. These datasets include document sets paired with four reference summaries. Pyramid evaluation is conducted on each reference summary by comparing it against the other three, with higher scores indicating greater alignment among the reference summaries and consistent content selection by different human summarizers.

This raises the question: Can we predict the agreement among \textit{human summarizers} using only the source document? To explore this, we averaged the Pyramid scores of all reference summaries for each document set to obtain $d_j^*$ and measured the correlation with PreSumm scores. As shown in Table \ref{tab:correlation_results_pyr_refSumm}, PreSumm outperforms all baselines in most datasets. This suggests that PreSumm can predict not only the performance of \textit{models} but also the likelihood of \textit{humans} agreeing on content selection using only the document. In other words, PreSumm can distinguish between documents with a clear main point that most people agree on and those lacking a central idea.

\section{Extrinsic Evaluation through Downstream Tasks}
\label{sec_applications}
In this section, we explore practical applications that benefit from predicting in advance the summarization model performance over documents. Additionally, these applications serve as extrinsic evaluations of our model. Specifically, we examine two use cases: (1) identifying in advance the documents where models perform poorly to enable manual summarization in hybrid  systems, and (2) filtering out noisy documents in a multi-document summarization setting.

\begin{table*}[t!]
    \begin{center}
   \resizebox{0.9\linewidth}{!}{

\begin{tabular}{l l c c c c} 
 \toprule
\multirow{2}{*}{Based on} & \multirow{2}{*}{Selection Method} & \multicolumn{ 2}{c}{BART} &  \multicolumn{ 2}{c}{Pegasus} \\ 
& & Replace 10\% & Replace 20\% & Replace 10\% & Replace 20\% \\ \hline 
 \multirow{ 4}{*}{Source} & Random & 0.425\textsuperscript{*} & 0.478\textsuperscript{*} & 0.394\textsuperscript{*} & 0.452\textsuperscript{*}\\
& Num. Entities & 0.437\textsuperscript{*}  & 0.508\textsuperscript{*}  & 0.416 & 0.486\textsuperscript{*}  \\
& Flesch & 0.427\textsuperscript{*}  &  0.494\textsuperscript{*}  & 0.403\textsuperscript{*}  & 0.472\textsuperscript{*} \\
  &PreSummReg& \textbf{0.451} & \textbf{0.525}& \textbf{0.423} & \textbf{0.499}\\ \midrule
\multirow{ 2}{*}{Source + Sys.} &Blanc &0.435\textsuperscript{*} & 0.512\textsuperscript{*} & 0.416 & 0.496\\
  &SummQa & 0.434\textsuperscript{*} & 0.502\textsuperscript{*} & 0.414 & 0.481\textsuperscript{*}\\

 \midrule
 \multirow{ 3}{*}{Sys.+ Ref.} & Rouge &  0.459 & 0.540  &  0.432 & 0.515\\   

  &Bert &  0.454 & 0.527 &  0.429 & 0.507\\
  &Meteor & 0.459 & 0.538 & 0.433 & 0.516\\

  \bottomrule
\end{tabular}
}
    \caption{Averaged ACU scores of system summaries after replacing selected summaries with manual summaries. Scores significantly worse than PreSummReg are marked with \textsuperscript{*}.} 
    \label{tab:filter_all}
\end{center}
\end{table*}

\subsection{Selecting Documents for Manual Summarization in Hybrid Systems}
\label{sec_task_filtering}
Here, we focus on a use case within a hybrid summarization system, where a fixed percentage of documents can be manually summarized within the available budget. Accordingly, we aimed to assess whether PreSummReg can effectively identify, in advance, documents that models are likely to fail on, allowing these documents to be prioritized for manual summarization. The goal is to maximize the overall score of the entire document set.

For this experiment, we used the generated summaries of two systems from our test set, Pegasus \citep{zhang2020pegasus} and BART \citep{lewis-etal-2020-bart}. For evaluation, we used the average of the human ACU scores, where instead of manually summarizing a selected document, we assigned it an ACU score of 1. For the manual summarization budget, we selected either 10\% or 20\% of the test set documents.
For statistical significance testing, we used the paired bootstrap test \citep{efron1994introduction} as explained in \citep{berg-kirkpatrick-etal-2012-empirical}. The detailed significance algorithm is provided in Algorithm 1 in the Appendix.

In addition to PreSummReg as a selection method, we evaluated several baseline methods. Baselines included Random selection and ranking methods based solely on the source document features, such as Flesch Reading Ease and the number of unique named entities. We also tested reference-free metrics such as Blanc \citep{vasilyev2020fill} and SummQA \citep{scialom2019answers}. However, a limitation of these reference-free metrics is that they require system-generated summaries, unlike PreSummReg. For comparison, we included reference-based metrics such as ROUGE-2 F1 \citep{lin2004rouge}, BERTScore F1 \citep{zhang2019bertscore}, and METEOR \citep{banerjee2005meteor}, which represent an upper bound since except the system summary they rely on the reference summary as well— an unavailable resource in real-world scenarios.

As shown in Table \ref{tab:filter_all}, PreSummReg significantly outperforms all source-only and reference-free baselines in most cases, and approach the upper bound set by reference-based metrics.
Overall, these experiments demonstrate that the PreSummReg model can effectively identify in advance documents that models are likely to fail on, optimizing the summarization process by saving time and resources.

\subsection{Multi-Document Summarization}
\label{sec_task_MDS}

In a Multi-document summarization (MDS) task, a set of documents on the same topic needs to be summarized. However, these sets often include noisy documents that can negatively impact model performance \citep{giorgi-etal-2023-open}. Additionally, summarizing a large number of documents poses challenges due to high input length, which can be costly or constrained by the token limits of certain models. Conventional MDS approaches typically concatenate all documents and truncate the input to meet the model’s token limit or user budget, leading to the exclusion of some documents. This raises the question: can we achieve better results by using PreSumm to identify and exclude noisy documents from the set?

To test this, we adopted the MDS MultiNews dataset \citep{fabbri-etal-2019-multi} and used the Pegasus and BART summarization models. We conducted summarization experiments with token limits of 256, 512, and 1024, where documents were truncated once the token limit was exceeded. In each experiment, we tested two variations: one where the documents were kept in their original order, and another where the documents were reordered based on their PreSummReg scores, with the lowest-scored documents placed at the end.
It is important to note that reordering the documents has an additional benefit—better document sequencing can enhance summarization quality even without excluding documents, as suggested by \citet{zhao-etal-2022-read}. Thus, we aimed to determine whether combining PreSummReg-based document exclusion with PreSummReg-based reordering would lead to improved summarization outcomes.


\begin{table}[t!]
    \begin{center}
        \resizebox{\columnwidth}{!}{
\begin{tabular}{c l c c c | c c c} \toprule & & \multicolumn{3}{c}{Pegasus}& \multicolumn{3}{c}{Bart} \\
\# Tokens & Order & R-1 & R-2 & R-L & R-1 & R-2 & R-L\\ [0.5ex] \midrule

\multirow{ 2}{*}{1024} & Original & 45.8 & 18.5 & 24.3 & 37.04 & 13.93 & 20.29 \\
& PreSumm & \textbf{46.1} & \textbf{18.9} & \textbf{24.6} & \textbf{37.34} & \textbf{14.27} & \textbf{20.52} \\ \midrule

\multirow{ 2}{*}{512} & Original & 43.9 & 17.2 & 23.5 & 35.58 & 12.99 & 19.58 \\ & PreSumm & \textbf{44.4} & \textbf{17.9} & \textbf{24.0} & \textbf{35.97} & \textbf{13.48} & \textbf{19.94} \\ \midrule

\multirow{ 2}{*}{256} & Original & 39.8 & 14.7 &21.5 & 33.51 & 11.79 & 18.57 \\ & PreSumm & \textbf{40.2} & \textbf{15.3} &\textbf{21.9} & \textbf{34.03} & \textbf{12.38} & \textbf{19.00} \\ \bottomrule
\end{tabular}
}
    \caption{ROUGE scores for multi-document summaries generated using different document ordering methods, truncated after a specified number of tokens.}
    \label{fig:mds_table}
\end{center}
\end{table}

As shown in Table \ref{fig:mds_table}, the PreSumm-ordered documents consistently achieve higher ROUGE scores compared to the original document order across all input length limits. PreSumm is significantly better across all metrics according to the Wilcoxon Rank Test \citep{Wilcoxon1945IndividualCB}, except for 1024-token limit with R-1 and R-L. This demonstrates that using PreSummReg to identify independent documents that models are likely to summarize unsuccessfully is also effective in enhancing multi-document summarization settings, as such independent documents contribute noise to the entire document set.

\section{Analysis}
\label{sec_analysis}
In this section, we aim to investigate PreSummReg to better understand the properties that make a document less likely to be successfully summarized by various summarization systems. To that end, we examine the influence of document-based features over PreSummReg by measuring correlation (Section \ref{subsec_analysis_correlations}). Then, we conducted a manual analysis to reveal more insights and explain the automatic results (Section \ref{subsec_manual_analysis}). 
Additional analysis examines how PreSumm deals with different corruptions can be found in Appendix  \ref{subsec_analysis_transformatiosn}.

\subsection{Document Feature Correlations}
\label{subsec_analysis_correlations}

To gain deeper insights into which document-based features most strongly influence the performance of summarization systems over a certain document, we analyzed the correlations between various document characteristics and the PreSummReg score. For features, we leveraged the baseline methods from Section \ref{subsec_baseline_models}, including the document statistics and the reading ease tests. In addition, we added the salient sentence location feature that uses the reference summary, and therefore it could not be used as a method in Section \ref{subsec_baseline_models}.

\paragraph{Salient Sentence Location.}
Previous work pointed that the salient information in news documents tend to be at the beginning of the document \citep{lebanoff2019scoreingpairs}. Since in this work we use mostly news domain datasets, we hypothesize that models may struggle to generate effective summaries when the main theme appears in the middle of the text or when multiple themes are present within the document. Therefore, we would like to set the location of the salient sentences in the document as a feature.
To determine the location of key sentences within a document, we adopted a method similar to \citep{nallapati2017summarunner, chen2018fastAbsSumm}, where each sentence in the document is ranked by its similarity to the reference summary, using ROUGE scores as the metric. We selected the top-5 or top-10 most salient sentences, and their positions were normalized by the total number of sentences in the document. The average of these normalized indices represents the typical location of the most important sentences in the document.

\begin{table}[t!]
    \begin{center}
\resizebox{0.8\columnwidth}{!}{
\begin{tabular}{l c } 
 \toprule
Feature & Correlation\\ [0.5ex] 
 \midrule

 Document length & -0.0956 \\ 
 Count of Numbers & -0.0576   \\
  \# of Unique Named Entities & -0.120  \\
   Flesch Reading Ease & 0.182  \\
 Flesch Kincaid Grade & -0.166 \\
  Avg Loc of Salient Sent (top 10) & -0.266  \\
   Avg Loc of Salient Sent (top 5)  & -0.305 \\
\bottomrule
\end{tabular}
}
    \caption{Pearson R correlations of document features to PreSummReg scores.}
    \label{fig:analysis_correlations_MERGED}

\end{center}
\end{table}

The correlation results are shown in Table \ref{fig:analysis_correlations_MERGED}. The most strongly correlated feature is the location of salient information. The negative correlation indicates that the earlier the key information appears in the document, the easier it is for the model to summarize. This observation aligns with expectations, as this is a well-known characteristic of the news domain, where essential information is typically presented at the beginning.

The Flesch Reading Ease and Flesch-Kincaid Grade Level scores show positive and negative correlations, respectively, indicating that more complex documents tend to result in poorer summarization model performance. In the same way, all basic document statistics show a relatively weak negative correlation, suggesting that greater document complexity—whether in length, the number of numerical values, or named entities—negatively impacts summarization performance.

Overall, while the correlation scores and their signs are consistent with our expectations, none of the features exhibited strong correlations. Consequently, we conducted a manual analysis to shed light on new document-based insights.

\subsection{Manual Analysis}
\label{subsec_manual_analysis}

To gain deeper insights into document performance, we conducted a manual analysis. Specifically, one of the authors read the 15 best and worst-ranked documents according to PreSummReg predictions. In general, this analysis found three unique properties of the bottom-ranked documents. 

\paragraph{Content Complexity.} The main and most common characteristic of low-ranked documents is that they often cover complex topics, such as science or politics, which include numerous numbers, intricate details, and long, difficult-to-follow sentences and documents. In contrast, the top-ranked documents were typically much shorter (some with only a single sentence), with simple words and topics.

\paragraph{Coherence.} Some bottom-ranked documents exhibited weak sentence-to-sentence connections or lacked sufficient background information, starting abruptly in the middle of a story, counting on specific terms and knowledge of a specific unique field. Sometimes it happens because of the crawling process of documents, that includes the image caption or sub-headers in the middle of the text. Such crawling issues were also seen in the top-ranked documents, but less frequently.

\paragraph{Theme Change.} Some low-ranked documents contain multiple, almost disjointed themes, making it difficult to determine which theme should be prioritized in the summary. This issue was especially pronounced in cases where the main theme was in the middle, requiring summarization models to go beyond its usual focus at the beginning of the document.

Overall, the low-ranked documents were notably more difficult to read, often requiring re-reading of certain sentences for comprehension, whereas the top-ranked documents were much more fluent and easier to follow. Examples of documents with these challenges are provided in Appendix \ref{sec_doc_example}.

To validate these findings, we conducted an extended annotation study with five NLP research students. They annotated a total of 50 documents (10\% of the test set), with each document being annotated by two students. Each annotation task involved a pair of documents: one from the 25 lowest-ranked by PreSummReg and another randomly selected document.
Following the insights from our preliminary manual analysis, annotators first identified independent flaws in each document, such as coherence issues or theme shifts. Then, they performed a comparative evaluation, determining which document contained more complex content and which was more fluent overall.

Due to the subjective nature of this task, there were disagreements over many annotation instances. To resolve these, we conducted a second round in which annotators discussed the disputed cases, attempting to reach a consensus. After this process, the inter-annotator agreement, measured by Cohen’s Kappa coefficient \citep{Cohen1960ACO}, reached 0.76 for Coherence, 0.66 for Theme-change, 0.92 for Complexity, and 0.53 for Fluency.

As can be seen in Table \ref{tab:manual_analysis}, our expanded analysis confirms the initial observation. Documents with lower PreSummReg scores are more prone to coherence issues and theme changes, are more complex, and are less fluent. This confirms that documents difficult for humans to read tend to be more challenging for summarization models as well.

\begin{table}[t!]
    \begin{center}
\resizebox{\columnwidth}{!}{
\begin{tabular}{l c c c c } 
 \toprule
& \multicolumn{2}{c}{Independent} & \multicolumn{2}{c}{Comparable} \\
\cmidrule(lr){2-3} \cmidrule(lr){4-5}  
Doc Type & Coherence & Theme & Complex & Fluent\\ [0.5ex] 
 \midrule
 Random & 28\%(44\%) & 12\%(12\%) & 32\%(36\%) & 60\%(84\%)\\
 Low PreSumm & 68\%(76\%) & 20\%(44\%) & 64\%(68\%) & 16\%(40\%)\\
 
\bottomrule
\end{tabular}
}
    \caption{Percentage of cases where all annotators (\textit{at least one annotator}) identified an \textit{Independent} flaw in a document or preferred one document on \textit{Comparable} criteria, categorized by document type.\protect\footnotemark}
    \label{tab:manual_analysis}

\end{center}
\end{table}

\footnotetext{The differences in percentages between the low PreSummReg documents and the random set are statistically significant in all cases according to paired bootstrap testing \citep{efron1994introduction}, except for Theme Change with full annotator agreement.}

These results also align with the correlation scores presented in Section \ref{subsec_analysis_correlations}. As noted, Reading Ease tests can indicate complexity, while the placement of salient information may relate to theme-changes and overall fluency. Although the correlation scores for these features were relatively weak compared to our manual analysis, this discrepancy may arise because our manual analysis focused on the lowest-scored documents rather than the full dataset. While this subset is more practically relevant, it also amplifies these properties, potentially leading to a stronger observed signal.

To compare this analysis to the actual performance rather than predicted by PreSummReg, we further examined documents with low average gold ACU scores. Interestingly, similar patterns emerged. However, some of these documents were penalized in ACU scoring due to misalignment between the reference summary and document content, rather than inherent flaws in the document itself. Since our model does not rely on reference summaries, it was unaffected by this misalignment and thus did not always rank these documents as low-quality.

\section{Conclusion}
\label{sec_conclusion}

In this work, we introduced PreSumm, a novel approach that opens up new research avenues in understanding the structural features that make a document less likely to be summarized successfully. Our findings suggest that documents that are more complex to read for humans are also ranked low by PreSumm models, implying the centrality of this feature for summarization models. We hope these insights will contribute to the design of more robust and effective models in the future.

\section*{Limitations}
\label{sec_ethical}
Our study covers the RoSE dataset extensively and focuses on summarization of the news domain only. Therefore, we cannot explore the complete space of summarization systems and we are limited to both the datasets and summarization systems that RoSE provides due to our extensive use of the manual ACU score. Because of this, some of the results could be due to other factors that relate to the dataset and would not generalize strongly outside of this study or to other domains.

Although we showed in our work that our model works relatively well on out-of-distribution data, we did not examine dataset out of the news domain. Therefore our conclusions are limited to this domain.

In future works, we would seek to train a similar model on a larger dataset. However, using ACU scores would be difficult because of the human labor involved, which could be avoided by using an annotated metric to train on. However, will make us biased towards the chosen metric, which is another limitation.

Out of distribution data is a major factor when it comes to model performance. To mitigate this would require greatly increasing the scope of the experiment and to train on a broader dataset for more accurate predictions.

Overall, overcoming these limitations would necessitate a much larger corpus with either a large set of automated or human annotated metrics to perform a similar study on a much larger set of the space of documents and summarization system.

\section*{Acknowledgments}
\label{sec_acknowledgments}
We thank the anonymous reviewers for their constructive comments. We would like to thank Lorenzo Flores, Sienna Hsu, and Cesare Spinoso-Di Piano for their annotations. This work was supported in
part by the IVADO Postdoctoral Fellowship and the Natural Sciences and Engineering Research Council of Canada.

\bibliography{bibliography}

\appendix
 \section{Appendix}
 \subsection{Implementation Details}
\label{subsec_Appendix_implement}
The SummEval models were built using the Longformer architecture, each with distinct output heads and training methodologies tailored to the task.

All models were trained for 5 epochs with a learning rate of 1e-6, using an 80\%-20\% train-validation split of the dataset. The base model used was the "allenai/longformer-base-4096" \cite{DBLP:journals/corr/abs-2004-05150}, configured to handle the maximum sequence length.

\subsubsection{Regression Head}
For regression tasks, a single linear layer was added to map the 768-dimensional CLS token embedding to a one-dimensional output, representing the predicted score.

\subsubsection{Classification Head}
For classification tasks, the Longformer backbone was paired with a more complex classification head. This head comprised a feedforward neural network with the following structure: a linear layer mapping 768 dimensions to 512, a ReLU activation, and a second linear layer reducing 512 dimensions to a single scalar output.
During the forward pass, the model computed individual scores from two heads, denoted as $s1$ and $s2$, and generated the final probability by subtracting these scores and applying the sigmoid function.

\section{Document-Based Transformation Analysis}

\label{subsec_analysis_transformatiosn}

To further investigate the influence of document-based features on model performance, we explored how the predicted score of a document changes when specific features are perturbed. By comparing the predicted scores before and after these transformations, we can gauge the importance PreSumm places on each feature. Intuitively, the transformations causing the largest change in predicted scores should indicate which features have the greatest impact on document performance in summarization. We applied these transformations to our test set. Below, we detail the transformations applied.
\begin{table}[t!]
    \begin{center}
    \resizebox{\columnwidth}{!}{

\begin{tabular}{l c c c } 
 \toprule
Transformation & Src. & 
 Trans. & Delta \\ [0.5ex] 
 \midrule
   Remove 10 salient sentences & 0.528 & 0.428 & -0.100 \\
 Keep last 3 sentences & 0.528 & 0.441 &-0.0877\\
 Delete 30\% of words & 0.528 &0.470 &-0.0584\\
Remove 5 salient sentences & 0.528  & 0.476 & -0.0523\\
Randomly Shuffle of Sentences & 0.528  & 0.486 & -0.0427\\
Move 10 salient sentences to end & 0.528&  0.502  & -0.0269\\
Keep first 3 sentences & 0.528 & 0.553 & 0.0246\\
 Remove first sentence & 0.528 & 0.506 & -0.0225 \\
  Move 5 salient sentences to end & 0.528   & 0.509 & -0.0199 \\
   Replace names w/ from bank & 0.528  & 0.512 & -0.0163 \\
 Replace names w/ spacy name & 0.528 & 0.518 & -0.0102 \\ 
   Corrupt Grammar & 0.528  & 0.520& -0.008\\
   Append contradictions & 0.528 &0.533 & 0.005\\
Delete 30\% of sentences & 0.528  &0.530&0.00150\\
\bottomrule
\end{tabular}
}
\caption{Predicted PreSummReg scores of documents before a transformation (Src.) and after (Trans.)}
    \label{fig:transformations}
\end{center}
\end{table}

\begin{description}
\item{\textbf{Removing content.}} We tested several removal strategies: removing the first sentence (often critical for summarization), removing 5 or 10 salient sentences (as defined in Section \ref{subsec_analysis_correlations}), and randomly removing 30\% of the words or sentences to disrupt fluency and coherence. Additionally, we removed all sentences except the first three or last three to assess the importance of content location.

\item{\textbf{Moving content.}} Given that salient information location was identified as a key feature in Section \ref{subsec_analysis_correlations}, we moved the salient sentences to the end of the document. We also randomly shuffled all sentences to disrupt coherence.

\item{\textbf{Replacing content.}} We replaced named entities to test the impact on consistency. We also introduced contradictions by adding negation sentences and corrupted the grammar by converting all verbs to their lemma forms.

\end{description}

As expected, the most impactful corruptions, with the highest changes in predicted scores, were removing the 10 most salient sentences and removing all content except for the last three sentences. Other significant transformations included deleting 30\% of the words, removing 5 salient sentences, and shuffling sentences randomly.

Surprisingly, moving the salient sentences to the end of the document had little effect, despite the location of salient information being one of the most influential features in Section \ref{subsec_analysis_correlations}.

Interestingly, content replacement, such as grammar corruption or adding contradictions, did not significantly affect PreSumm's performance. It also appears that strong perturbations, such as deleting 30\% of sentences, did not lead to large differences in scores. This might be because these artificial corruptions are not natural and therefore deviate too much from the patterns seen during model training, causing the model to mispredict their impact on summarization performance.

\begin{algorithm}
\caption{PreSumm vs method X Paired Bootstrap Significance Test}

\begin{algorithmic}[1]
\label{algo_significance}
\State \textbf{Input:} Test set of documents and system summaries of a specific system, PreSumm scores, X scores, ACU scores
\State \textbf{Output:} p-value

\State Extract the current difference in performance between PreSumm filtering and X filtering, denoted as \texttt{original\_diff}.
\State Initialize \( s = 0, b=10,000 \)
\State $n$ is the number of instances (documents)

\For{$i \gets 1$ to $b$}
    \State \parbox[t]{\dimexpr\linewidth-\algorithmicindent}{ Sample n instances with replacement from the test set\strut}
    
    \State \parbox[t]{\dimexpr\linewidth-\algorithmicindent}{Replace 10\% or 20\% of system summaries with reference summaries according to PreSumm, and average the ACU scores to get \texttt{PreSumm\_filter\_score}\strut}
    \State \parbox[t]{\dimexpr\linewidth-\algorithmicindent}{Replace 10\% or 20\% of system summaries with reference summaries according to X, and average the ACU scores to get \texttt{X\_filter\_score}\strut}
  \If {PreSumm\_filter\_score - X\_filter\_score >    2 $\times$ \texttt{original\_diff}}
    \State  \( s = s + 1 \)
   \EndIf
\EndFor

\State Compute \( p\_val = \frac{s}{b} \)

\end{algorithmic}

\end{algorithm}

\section{Annotation Guidelines}
Below are the annotation guidelines provided to annotators for the manual analysis described in Section \ref{subsec_manual_analysis}.

Read the following documents carefully and answer the questions below for each.

\textbf{Document A:}

<document A>

\begin{itemize}
\item Did you notice any coherence issues in Document A?
Examples: corruptions, unrelated sentences, topics starting abruptly without enough background.

Yes/No

\item Does Document A contain a theme change?
Examples: discusses topics that deviate significantly from the main theme, or there are multiple main themes.

Yes/No

\end{itemize}

\textbf{Document B:}

<document B>

\begin{itemize}
\item Did you notice any coherence issues in Document B?
Examples: corruptions, unrelated sentences, topics starting abruptly without enough background.

Yes/No

\item Does Document B contain a theme change?
Examples: discusses topics that deviate significantly from the main theme, or there are multiple main themes.

Yes/No

\end{itemize}

\textbf{Comparison Questions:}

\begin{itemize}
\item Which document had more complex content?

Document A/Document B

\item Which document is more fluent to read?

Document A/Document B
\end{itemize}

\section{Example Documents}
\label{sec_doc_example}

Tables \ref{tab:doc_example} and \ref{tab:doc_example2} present examples of documents from the 15 lowest PreSummReg scores, annotated with their respective challenges. For comparison, Table \ref{tab:doc_example3} provides examples of documents from the 15 highest PreSummReg scores.

\begin{table*}[t]
    \centering
    \begin{tabular}{|p{0.2\linewidth}|p{0.75\linewidth}|}
\hline
\textbf{Challenging Characteristics} & \textbf{Document} \\ \hline

Coherence & \texttt{Judge Thokozile Masipa did the same for the lawyers on Thursday, urging them to make good use of the upcoming fortnight break for the Easter holidays.\newline In that spirit, here are a few questions that have been niggling me in recent days.\newline Tweet your thoughts and suggestions to @BBCAndrewH. I will be taking a week off and then focusing on South Africa's general election before returning to the hard benches of Courtroom GD on 5 May.} \\ \hline

Theme Change & \texttt{The images, taken by Syd Shelton, from Pontefract, include pictures of The Clash, Misty in Roots and The Specials.\newline The collection also features photos taken at the Rock Against Racism Carnival at Victoria Park, Hackney, which attracted a crowd of 100,000.\newline The show runs from Friday to 3 September at the Impression Gallery.\newline The Rock Against Racism (RAR) movement formed in response to controversial remarks made by Eric Clapton in 1976.\newline In the following years, RAR staged marches, festivals and more than 500 concerts in the UK in a bid to fight racism through music.\newline Shelton, who studied Fine Art in Leeds and Wakefield, said he became involved with the movement after returning to the UK from America in 1976.\newline He said: \"I was appalled at the state of race relations in Britain, in particular things like the Black and White Minstrel Show and the signs I saw in some windows saying 'No Blacks, No Dogs, No Irish'.\newline \"It was a pretty serious situation and I always loved music and very quickly hooked up with the people that had set up RAR.\newline \"It was a bizarre mixture of people, photographers, graphic designers, writers, actors and, of course, musicians.\newline \"We were very lucky in the sense that we tuned in to that explosion of punk and UK reggae and brought the two together. That said more about what RAR was about than any of the slogans we may have shouted from the stage.\"\newline He added: \"I hope the exhibition shows that you can change things and you can actually take a stand, even in the most difficult of situations.\newline ...} \\ \hline

\end{tabular}
\caption{Example documents from the 15 lowest predicted scores by PreSummReg, categorized by their challenging characteristics.}
\label{tab:doc_example}
\end{table*}
\begin{table*}[t]
    \centering
    \begin{tabular}{|p{0.2\linewidth}|p{0.75\linewidth}|}
\hline
\textbf{Challenging Characteristics} & \textbf{Document} \\ \hline

Content Complexity & \texttt{Welsh language minister Alun Davies told AMs it would help efforts to reach that goal stay on the right track.\newline Targets to meet growing demand for Welsh-speaking teachers and public sector workers will also be set.\newline Culture committee chairwoman Bethan Jenkins said AMs had been told 70\% more Welsh-medium teachers were needed.\newline Mr Davies responded that around a third of teachers in Wales could speak Welsh, and that the challenge was to see if more of them would be willing to teach through the medium of Welsh.\newline Earlier this month, Welsh language commissioner Meri Huws called for \"radical change\" in the education system to ensure all children under the age of seven were \"immersed\" in Welsh.":} \\ \hline

Content Complexity & \texttt{He said new forests would slow flooding by trapping water with their roots.\newline The idea of \"rewilding\" the uplands is catching on fast as parts of Britain face repeated flooding, with more rainfall on the way.\newline Environment Secretary Owen Paterson said he would seriously consider innovative solutions like rewilding.\newline The government has been criticised for being slow to capitalise on the benefits of capturing rain where it falls.\newline Lord Rooker, a Labour peer, said too much emphasis had been attached to the look of the countryside rather than practical considerations like trapping water.\newline \"We pay the farmers to grub up the trees and hedges; we pay them to plant the hills with pretty grass and sheep to maintain the chocolate box image, and then wonder why we've got floods,\" he said.\newline The idea of reintroducing forests into catchments has been strongly supported by several leading scientists.\newline The government is sponsoring a handful of catchment trials to assess the potential of the upstream areas to catch water and send it slowly downhill. \newline ...} \\ \hline

\end{tabular}
\caption{Example documents from the 15 lowest predicted scores by PreSummReg, categorized by their challenging characteristics.}
\label{tab:doc_example2}
\end{table*}
\begin{table*}[t]
    \centering
    \begin{tabular}{|p{1\linewidth}|}
\hline
\textbf{Document} \\ \hline

\texttt{Eve: Where are we meeting?\newline Charlie: at the entrance\newline Nicole: yes, it's the best place. We would't find each other inside, it'll be too crowded\newline Eve: ok!} \\ \hline
 \texttt{Jair: Still busy?\newline Callum: Yes a little sorry\newline Jair: ok} \\ \hline
  \texttt{A 16-year-old girl is anxiously awaiting blood test results after sitting on a needle on a bus. Francesca Palmer-Norris was on the top deck of the number 24 Brighton and Hove Bus Company vehicle when she was pricked by the needle. The worried student, from Brighton, East Sussex spent the next four hours in hospital where she was given a hepatitis jab and had blood tests. Worried:\u00a0Francesca Palmer-Norris is awaiting blood test results after sitting on a needle on the top deck of the bus . Speaking about the incident, Ms Palmer-Norris said: 'My friend and I had got on the bus to go home and we were sat on the top. 'I suddenly had this shooting pain in the back of my leg. I reached down and pulled out a needle that had snapped in half. 'Then I looked down the side of the bus seat and there were packets and a syringe on the floor and the rest of the needle.' She added: 'When the bus reached the next stop, I explained to the driver what had happened and he said it was best to go to the hospital.' She was given a jab and had blood tests before going home that night. Francesca Palmer-Norris was on the top deck of the number 24 Brighton and Hove Bus Company vehicle when she was pricked by the needle (stock image) Ms Palmer-Norris said: 'The worrying thing now is I am waiting for the results to come back. 'My head is all over the place - I can't sleep.' The bus company said the driver closed the top deck of the bus after the incident and took the vehicle for a full inspection as 'soon as practicably possible'. Adrian Tullett, head of operations at Brighton and Hove Bus Company, said the incident was being investigated using CCTV footage. He added: 'The driver followed procedure and secured off the top deck as soon as he was made aware of an object that needed removing from the seating area. 'He took the vehicle out of service for a full inspection as soon as was practically possible. 'We would like to reassure passengers we take these matters very seriously and that all our buses get a visual inspection at the end of each journey. Our customer services team is liaising direct with the girl's family.' Sussex Police is also investigating the incident.} \\ \hline

\end{tabular}
\caption{Example documents from the 15 highest predicted scores by PreSummReg.}
\label{tab:doc_example3}
\end{table*}
 
\end{document}